\documentclass{article}

\usepackage{PRIMEarxiv}
\usepackage{microtype}
\usepackage{graphicx}
\usepackage{subfigure}
\usepackage{floatrow}
\usepackage{amssymb}
\usepackage{mathtools}
\usepackage{multirow,tabularx}
\usepackage{dblfloatfix} 
\usepackage{colortbl} 
\usepackage{breqn} 
\usepackage{amsmath} 
\usepackage{dblfloatfix}
\usepackage{enumitem}
\usepackage{booktabs}
\usepackage{threeparttable}
\usepackage{wrapfig,lipsum,booktabs}
\usepackage{makecell} 
\newfloatcommand{capbtabbox}{table}[][\FBwidth]
\usepackage[utf8]{inputenc} 
\usepackage[T1]{fontenc}    
\usepackage{hyperref}       
\usepackage{url}            
\usepackage{booktabs}       
\usepackage{amsfonts}       
\usepackage{nicefrac}       
\usepackage{microtype}      
\usepackage{xcolor}         
\usepackage[numbers]{natbib}

\title{Deep Metric Loss for Multimodal Learning
} 

\author{%
	Sehwan Moon  \\
	Electrical Engineering and Computer Science\\
	Gwangju Institute of Science and Technology\\
	\texttt{sehwanmoon@gist.ac.kr} 
  \And
        Hyunju Lee \\
        Electrical Engineering and Computer Science\\
        Gwangju Institute of Science and Technology\\
        \texttt{hyunjulee@gist.ac.kr} 
}

\begin{document}

	\maketitle

	\begin{abstract}
		Multimodal learning often outperforms its unimodal counterparts by exploiting unimodal contributions and cross-modal interactions. However, focusing only on integrating multimodal features into a unified comprehensive representation overlooks the unimodal characteristics. In real data, the contributions of modalities can vary from instance to instance, and they often reinforce or conflict with each other. In this study, we introduce a novel \text{MultiModal} loss paradigm for multimodal learning, 
which subgroups instances according to their unimodal contributions. \text{MultiModal} loss can prevent inefficient learning caused by overfitting and efficiently optimize multimodal models.
On synthetic data, \text{MultiModal} loss demonstrates improved classification performance by subgrouping difficult instances within certain modalities. On four real multimodal datasets, our loss is empirically shown to improve the performance of recent models. Ablation studies verify the effectiveness of our loss. Additionally, we show that our loss generates a reliable prediction score for each modality, which is essential for subgrouping. Our \text{MultiModal} loss is  a novel loss function to subgroup instances according to the contribution of modalities in multimodal learning and is applicable to a variety of multimodal models with unimodal decisions. Our code is available at https://github.com/SehwanMoon/MultiModalLoss.
	\end{abstract}

\section{Introduction}

Multimodal learning has received considerable attention from academia and industry owing to its progress in a wide range of real-world applications. The fusion of modalities is one of the most important and popular topics in multimodal learning research. There are three basic types of modality fusion: early (feature-level), intermediate (learning the representation of each modality and unifying multiple heterogeneous representations into a common latent space), and late (decision-level)  \citep{atrey2010multimodal, ramachandram2017deep}. 

Among these, most studies have adopted intermediate approaches owing to their flexibility \citep{ramachandram2017deep}. This requires carefully considering how and when to fuse modalities to optimize the objective function. In particular, most existing studies have shared representation mapping by considering unimodal information only when generating latent variables \citep{kamath2021mdetr,bendre2021generalized,gu2023dual}. The cases where all instances are equally influenced by each modality do not present any problems. However, with real-world data, this equality cannot be guaranteed. In several cases, the different modalities yield conflicting or predominant results (\textit{e.g.}, diagnoses should be determined by different weights of combinations of factors, including imaging, clinical presentation, and histological examination, depending on the patient's individual case).

\begin{figure}
	\begin{center}
		\centerline{\includegraphics[width=0.65\textwidth]{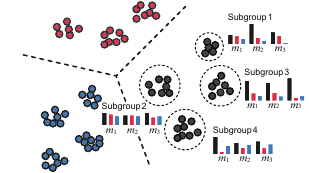}}
		\vskip 0.05in
		\caption{Illustration of the proposed \text{MultiModal} loss. It generates multiple subgroups based on the classification contributions and difficulty of each modality. Circles with different colors represent instances of different classes. Each bar represents the probability for each class with the same color on three modalities, $m_1$, $m_2$, and $m_3$.
			Subgroup 1 is largely contributed by $m_2$, subgroup 2 is difficult to classify for all modalities, and subgroup 3 is easy on all modalities. Subgroup 4 is largely contributed by $m_1$ and has a higher probability for the blue class than the red class.}
		\label{fig1}
	\end{center} 
	\vskip -0.3in
\end{figure}

Recently, considering unimodal decisions has become popular among intermediate fusion methods because: 1) they can use the existing pretrained or state-of-the-art unimodal models, 2) they preserve unimodal decisions, 3) they are useful in considering the shared cross-modal interactions rather than mapping shared representations when a huge unimodal for each modality is needed	\citep{ghosal2018contextual,xu2018co,wei2019neural,dong2019mlw,tsai2019multimodal,su2020msaf,joze2020mmtm,siddiqui2020multimodal,delbrouck2020transformer,sood2021multimodal,saad2021learning,chen2021multimodal,moon2022moma,9956592,wang2022fmfn,wang2022energy,yin2022x}. However, numerous researchers only apply summation or a fully connected layer after concatenation to reach the final decision, which may lead to decision loss or overfitting due to insufficient amount of data, or inefficient equal backpropagation of modalities \citep{liu2018learn}.

To improve this situation, we proposed a novel \text{MultiModal} loss that subgroups instances using unimodal contributions, as shown in Figure~\ref{fig1}. The new loss is a proxy-based loss that subgroups instances based on their modality contributions. 
To make reliable predictions for each modality, we used a soft attention mechanism.
\text{MultiModal} loss can be easily applied to fusion models considering unimodal decisions. 

{\bf Contributions.} (1) We proposed a new loss for subgrouping according to modality contribution in multimodal learning, which can be applied to a variety of multimodal models (Sec. 3). (2) We generated synthetic data containing difficult instances in certain modalities and demonstrated that prior approaches ignored difficult modalities, which degraded the final performance. We demonstrated that our subgrouping approach performs well without ignoring difficult modalities (Sec. 4). (3) We applied our loss to recent multimodal models on four real-world datasets to demonstrate performance improvements (Sec. 4). (4) We presented an analysis to understand and validate the design decisions of our loss (Sec. 5).

\section{Related Works}
\subsection{Multimodal learning}
Multimodal learning is gaining increasing attention as the amount of multimodal data with rich correlations increases. Numerous studies use information from multiple modalities to extend the boundaries of unimodal frameworks for specific tasks, such as action recognition \citep{wu2021spatiotemporal, li2020sgm}, audiovisual speech recognition \citep{kashevnik2021multimodal, song2022multimodal}, and diagnostic classification \citep{eddy2020integrated, moon2022sdgcca}. For efficient multimodal learning, Mulmix \citep{liu2018learn} multiplicatively combines information from multimodal data to focus on more reliable modalities and ignore weak modalities. However, this method has limitations in that it ignores difficult modalities and is sensitive to outliers and unseen data.

\subsection{Deep Metric Learning}
The aim of metric learning is to measure the similarity of instances using an optimal distance metric. Deep metric learning utilizes deep neural network to learn the nonlinear embeddings. Contrastive loss \citep{bromley1993signature, chopra2005learning} and triplet loss \citep{ weinberger2009distance, wang2014learning} are the most prominent loss functions used for this purpose. Contrastive loss takes a pair of embeddings and minimizes their distance if they belong to the same class. Otherwise, the distance is maximized. Triplet loss sets an anchor reference in the embedding space, associates it with   positive and negative instances, and forces the distance between the anchor and the positive pair to be smaller than the distance between the anchor and the negative pair. However, as the batch size increases, the number of pairs and triplets grows exponentially, thus leading to resource consumption problems.
\newpage
\subsection{Proxy-Based Loss}
\begin{wrapfigure}[26]{r}{0.3\linewidth}
	\begin{center}
		\vskip -0.85in
		\centerline{\includegraphics[width=1.04\textwidth]{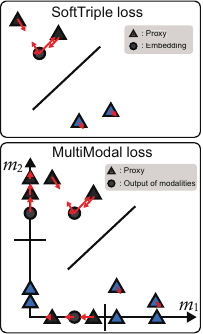}}
		\vskip -0.27in
		\caption{Illustration of the differences between SoftTriple and MultiModal loss. SoftTriple minimizes the distances between instances and proxies in the same class based on distance (\textit{i.e.} similarity). Otherwise, the distance is maximized. MultiModal loss additionally considers and preserves the characteristics of each modality ($m_1$ and $m_2$).}
		\label{fig2}
	\end{center}
\end{wrapfigure}
Recently, proxy-based losses have been designed to address this increasing complexity by introducing proxies as artificial data points. Proxy-NCA loss \citep{movshovitz2017no} assigns each class a single proxy, which is a trainable parameter in the embedding space. Proxy-NCA loss associates each instance with proxies. ProxyAnchor loss \citep{kim2020proxy} considers the fine-grained relationship between entire instances in the batch and the proxy.

SoftTriple loss \citep{qian2019softtriple} assigns multiple proxies to each class to reflect the intra-class variance found in real-world data. This loss is defined by measuring the similarity between each instance and the proxies of all classes, as follows.
\begin{equation}
	\ell_{\text{SoftTriple}}  = -log{  exp(\lambda(S_{i,y_i}-\delta))  \over exp(\lambda\left(S_{i,y_i} - \delta  \right) )+ \sum_{j\neq y_i} exp(\lambda S_{i,j})},
\end{equation}
\begin{equation}
	S_{i,c} = \sum_{k}{{exp({1 \over \gamma } x_{i}^{\top} w_c^k) \over \sum_{k}exp({1 \over \gamma } x_i^{\top} w_c^k)} x_i^{\top} w_c^k},
\end{equation}
where $x_i$ denotes the embedding of the $i$th instance, $y_i$ is the corresponding label, $ w_c^k$ are trainable parameters representing proxy $k$ of class $c$, $\lambda$ denotes the scaling factor, $\delta$ is the margin between the instance and proxies from different classes, and $\gamma$ is the entropy regularizing factor.

Our \text{MultiModal} loss is inspired by proxy-based loss and subgroups instances based on their unimodal contribution. However, the existing proxy-based loss has the limitation that it cannot preserve the characteristics of each modality (\textit{i.e.}, output of unimodal). Reliable output of each modality is essential for subgrouping. We are the first to address the above limitation and solve the multimodal subgrouping problem.

\section{\text{MultiModal} Loss}

Our \text{MultiModal} loss takes multiple proxies as the center of the subgroup based on the contribution of each modality. Reliable outputs must be maintained for each modality, otherwise the model will be biased toward the easier modalities, essentially neglecting the harder ones and ignoring their unique characteristics. Our loss trains the model to subgroup instances while also classifying within each modality. Figure~\ref{fig2} illustrates the differences between SoftTriple loss and the proposed losses. The gradients for proxy and instance for \text{MultiModal} loss and SoftTriple loss can be found in Appendix A. Key activities include taking multiple proxies as the center of the subgroup based on the contribution of each modality, and using soft attention to obtain high entropy prediction scores for each modality.

We denote the output (prediction) of the last layer of the $i$-th  instance for the $m$-th modality as $x_i^{m}\in \mathbb{R}^{C}$ and the corresponding label as $y_i$. $[w_c^{1,m},...,w_c^{K,m}] \in \mathbb{R}^{C \times K}$ is a proxy for class $c$, $K$ denotes the number of proxies, and $C$ is the number of classes.

\begin{wrapfigure}[21]{r}{0.4\linewidth}
	\begin{center}
		\vskip - 0.5in
		\centerline{\includegraphics[width=1.2\textwidth]{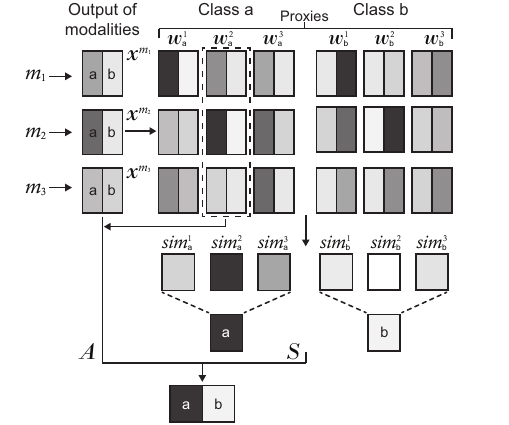}}
		\vskip -0.26in
		\caption{Architecture of  \text{MultiModal} loss. The darkness of the color represents the probability.
			The similarity $sim$ between the output\boldsymbol{$ $x} (as query) and the proxies\boldsymbol{$ $w} (as key) is determined, and the proxies are highlighted based on their similarity score. The predicted distribution over different classes is computed with the attended output\boldsymbol{$ $A} (sum of existing outputs and highlighted proxies) and the normalized similarities\boldsymbol{$ $S} obtained from each class.}
		\label{fig22}
	\end{center}
\end{wrapfigure}

Figure~\ref{fig22} illustrates the architecture of \text{MultiModal} loss.
Proxy similarity scores are determined by the dot product of the query (outputs with different modalities) with all keys (proxies):
\begin{equation}
	sim_{i,c}^k = \sum_{m}{{x_{i}^{m}} \cdot w_c^{k,m}}, 
\end{equation}
where $\cdot$ denotes the dot product.
The attention weights are normalized over all proxies as follows.
\begin{equation}
	att_{i,c}^k = {{exp\left({1 \over \gamma }sim_{i,c}^k \right) \over \sum_{c}\sum_{k}exp\left({1 \over \gamma } sim_{i,c}^k\right)}},   
\end{equation}
where $\gamma$ is the scaling factor. To effectively adjust the output of each modality, we take the weight sum of the proxies according to the attention weights, $att_{i,c}^k$, which are calculated by the soft attention. Subsequently, the attended output is calculated by 
\begin{equation}
	A_i = \sum_{m}{(\sum_{c}{\sum_{k}att_{i,c}^k w_c^{k,m}+ J)}}\circ x_{i}^{m},
\end{equation}
where $A_i\in\mathbb{R}^{C}$, $\circ$ denotes the Hadamard product or element-wise multiplication, $J$ denotes a matrix of ones.
The proxies of other classes may be unrelated in SoftTriple loss \citep{qian2019softtriple}. However, in our problem, in some cases, the classes differ while playing the same role (\textit{e.g.}, 
i.e., where only one modality prevails or all modalities are difficult).
To induce proxies corresponding to each class, we normalized based on class.
The normalized similarity score between the outputs of modalities and the class $c$ is defined as 
\begin{equation}
	S_{i,c}= \sum_{k} {{{exp\left({1 \over \gamma } sim_{i,c}^k\right) \over \sum_{c}exp\left({1 \over \gamma } sim_{i,c}^k\right)}} sim_{i,c}^k}.
\end{equation}
By applying the normalized similarity score and the attended output, we define the \text{MultiModal} loss as 
\begin{equation}
	\ell_{\text{MultiModal}}  = -log{exp(S_{i,y_i}+A_{i,y_i})  \over \sum_{c} exp(S_{i,c}+A_{i,c})},
\end{equation}
where $A_{i,c}$ denotes the attended output of the class $c$.

\section{Experiments}
In this section, we describe our experiments. First, a synthetic dataset was used to validate the performance of \text{MultiModal} loss. Next, we applied our loss to recent multimodal models on four real-world datasets to measure its effectiveness: the Ryerson Audio-Visual Database of Emotional Speech and Song (RAVDESS) dataset \citep{livingstone2018ryerson}, OPPORTUNITY recognition dataset \citep{chavarriaga2013opportunity}, EPIC-KITCHENS \citep{damen2020epic}, 
The Cancer Genome Atlas (TCGA) \citep{weinstein2013cancer}.

We compared existing fusion methods, namely cross entropy loss after summation (\text{SUM + CE}), cross entropy loss after a neural network layer (\text{NN + CE}), and Mulmix \citep{liu2018learn}. Additionally, we compared the two proxy-based losses, SoftTriple and ProxyAnchor loss.
We set the number of proxies to 20 and $\gamma$ to 0.1 for \text{MultiModal} loss as SoftTriple loss exhibited no effect on performance with a sufficiently large number of proxies \citep{qian2019softtriple}. We used the default hyperparameters experimented in the original paper for Mulmix, SoftTriple (the number of centers is 20), and ProxyAnchor.

In addition to the metrics of the base models, we considered appropriate metrics, namely accuracy (ACC), macro F1 (FM), weighted F1 (FW), and the Matthews correlation coefficient (MCC), for imbalanced data.

All experiments were conducted using an Nvidia Titan RTX GPU in an Ubuntu v.18.04 environment using Python v.3.7.9 and PyTorch v.1.9.0. The results were averaged over five runs using a random seed in [0--4]. For five-fold cross-validation (CV), we fixed the random seed to zero.

\subsection{Synthetic Data}
We investigated the effect of subgrouping difficult instances on some modalities. We generated synthetic data that included difficult instances on some modalities by adding noise.

{\bf Data Generation.} 
We generated three modalities for the synthetic data, denoted by $m_1$, $m_2$, and $m_3$. We generated $m_i \sim \mathcal{N}(0,\,1), i=1,2,3$, where $m_i$ is a 2,000-dimensional random vector. $y_i$ was generated as follows.
\begin{equation}
	y_i = {exp\left( ( m_i+\sigma_i\cdot e )\beta_i + \epsilon \right) \over \ 1 + exp\left(  ( m_i+\sigma_i\cdot e )\beta_i + \epsilon  \right),}
\end{equation}
\indent where $\beta_i$ represents the coefficient vector, $e$, $\epsilon$ represents the random error following $ \mathcal{N}(0,\,1)$, and $\sigma_i$ denotes the noise control parameter. The coefficient vectors, $\beta_i$, were set as sparse with 20 non-zero coefficients from $ \mathcal{N}(0,\,2)$. We set $\sigma_i = 0.2$ for the easy modality and $\sigma_i = 10$ for the noisy modality containing difficult instances. Subsequently, we generated two-class training, validation, and test sets of 10,000, 1,000, and 10,000 data items, respectively.

{\bf Training Setting.}
For each modality, we considered a four-layer perceptron with an input size of 2,000 and hidden layer sizes of 500, 100, 20 and 2 for each layer, using SoftMax for the last layer. We compared the late fusion performance of the three modalities. We applied the Adam \citep{kingma2014adam} optimizer with a weight decay of 1e-4 and an early stopping patience of 100. We tuned the learning rate to \{5e-3, 5e-4, 5e-5, 5e-6\} on the validation set (see Appendix B for our tuning results).

\begin{table}[t]
	\vskip -0.05in
	\caption{Comparison of the average classification accuracy between different multimodal situations on synthetic data.}
	\label{table1}
	\begin{center}
		\begin{small}
			\centering
			\setlength\tabcolsep{2.5 pt}
			{\renewcommand{\arraystretch}{1.2}
				\begin{tabular}{cccc|cccccc}
					\toprule
					\multirow{2}{*}{{\begin{tabular}[c]{@{}c@{}}\# of Noise \\ Modality\end{tabular}}} & \multicolumn{3}{c|}{{Unimodal}}  & \multirow{2}{*}{{\begin{tabular}[c]{@{}c@{}}SUM\\+ CE\end{tabular}}} & \multirow{2}{*}{{\begin{tabular}[c]{@{}c@{}}NN\\+ CE\end{tabular}}} & \multirow{2}{*}{{\begin{tabular}[c]{@{}c@{}}Mul\\mix\end{tabular}}}  & 
					\multirow{2}{*}{{\begin{tabular}[c]{@{}c@{}}{Soft}\\{Triple}\end{tabular}}}  & 
					\multirow{2}{*}{{\begin{tabular}[c]{@{}c@{}}{Proxy}\\{Anchor}\end{tabular}}}  & 
					\multirow{2}{*}{{\begin{tabular}[c]{@{}c@{}}{Multi}\\{Modal}\end{tabular}}}  \\ \cline{2-4}
					& \boldsymbol{$ $m_1} & \boldsymbol{$ $m_2} & \boldsymbol{$ $m_3}  \\ \midrule 
					0                                                                        & 86.1\tiny{$\pm$0.1}      & 87.1\tiny{$\pm$0.1}     & 86.1\tiny{$\pm$0.3}     & 93.1\tiny{$\pm$0.2}                                                      & \underline{94.3\tiny{$\pm$0.2}}                             & 92.7\tiny{$\pm$0.6}   & 77.6\tiny{$\pm$16.2}      & 81.4\tiny{$\pm$6.7}                 & \textbf{96.0\tiny{$\pm$1.8}}      \\
					1 (\boldsymbol{$ $m_3})                                                                & 86.2\tiny{$\pm$0.2}     & \underline{87.4\tiny{$\pm$0.1}}      & 55.9\tiny{$\pm$0.1}     & 85.3\tiny{$\pm$0.4}                                                     & 87.2\tiny{$\pm$0.4}                                           & 84.8\tiny{$\pm$0.4}   & 74.0\tiny{$\pm$16.8}  & 75.1\tiny{$\pm$14.3}           & \textbf{92.1\tiny{$\pm$2.1}}      \\
					2  (\boldsymbol{$ $m_2,m_3})                                                                 & \underline{86.1\tiny{$\pm$0.2}}      & 53.7\tiny{$\pm$0.1}     & 54.0\tiny{$\pm$0.1}      & 68.6\tiny{$\pm$1.2}                                                     & 80.0\tiny{$\pm$2.6}                           & 68.9\tiny{$\pm$1.1}         & 64.7\tiny{$\pm$9.8}    & 64.1\tiny{$\pm$12.9}          & \textbf{86.4\tiny{$\pm$0.7}}     \\\bottomrule
				\end{tabular}
			}
		\end{small}
	\end{center}
\end{table}
\begin{figure}
	\vskip -0.14in
	\centerline{\includegraphics[width=\columnwidth]{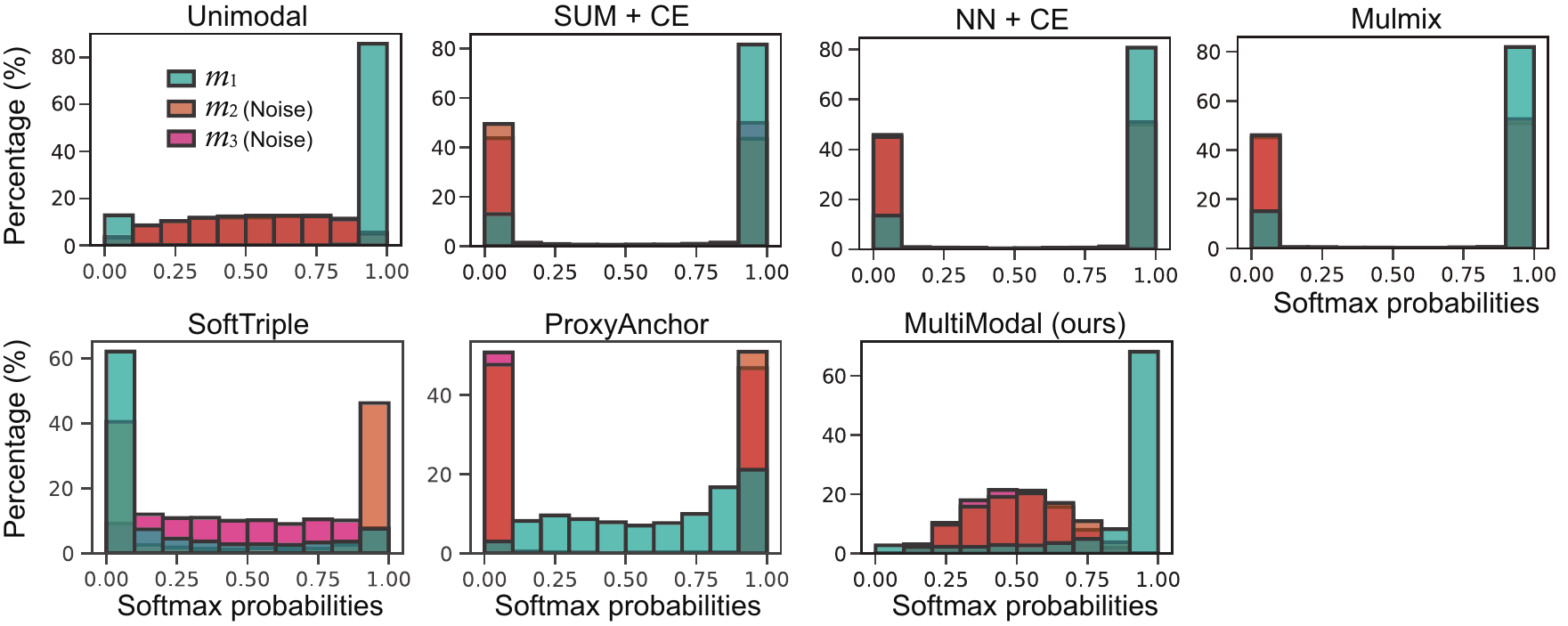}}
	\vskip -0.0in
	\caption{Percentage histogram of the predicted classification probability conditioned on true labels for each modality on synthetic data with two noise modalities.}		\label{fig3}
	\vskip -0.05in
\end{figure}
{\bf Results.} 
Table~\ref{table1} lists the average classification accuracies for the number of noise modalities. When the data had no noise modality, all methods using multiple modalities outperformed the unimodal baseline. After introducing noise, the existing fusion methods (\textit{i.e.}, \text{SUM + CE}, \text{NN + CE}, \text{Mulmix}), SoftTriple loss, and ProxyAnchor loss exhibited worse performance compared with the unimodal model. However, by benefiting from the subgrouping of difficult instances in certain modalities, the proposed method retained awareness of the most important modalities and achieved competitive results.

A percentage histogram of the predicted classification probabilities is shown in Figure~\ref{fig3}. The traditional fusion methods produced highly confident predictions (low entropy) for which the unimodal prediction value was corrupted. SoftTriple and ProxyAnchor loss misrepresented the noise modalities, thus indicating their inability to preserve the characteristics of each modality. However, the proposed loss function provided reliable results (high entropy) while preserving the valuable unimodal information.

\subsection{RAVDESS Dataset}

First, we applied our loss function to the state-of-the-art Intermediate Attention fusion model \citep{9956592} on RAVDESS. RAVDESS contains multimodal video and audio data for emotion recognition tasks. Seven emotional classes, namely, angry, calm, disgust, fear, happy, sad, and surprise, were built from 1,440 videos of 24 professional actors (12 females and 12 males). We applied the hyperparameters of the original model, and because the five-fold CV setting of the Intermediate Attention fusion model is ambiguous, we borrowed the five-fold stratified CV setting from a recent paper \citep{luna2021proposal}. We included the validation set for training and used the early stopping with a patience of ten epochs (monitoring training loss as a criterion).

{\bf Results.} 
The results of applying our loss to the Intermediate Attention fusion model on RAVDESS are presented in Table~\ref{table2}. Evidently, our \text{MultiModal} loss improved performance over the baseline Intermediate Attention fusion model using the NN + CE method. By contrast, proxy-based losses, including SoftTriple and ProxyAnchor, exhibited very poor performance.

\begin{table*}[t]
	\vskip -0.1in
	\caption{Result on RAVDESS for emotion recognition (7 classes).}\label{table2}	
	\vskip -0.1 in
	\begin{center}
		\begin{small}
			\centering
			\setlength\tabcolsep{4.5 pt}
			{\renewcommand{\arraystretch}{1.2}
				\begin{tabular}{lcccccc}
					\toprule
					\multicolumn{1}{l}{{Model}}& {Modality} & {Method}    & {ACC}   & {FM} & {FW} & {MCC}    \\ \midrule MULT \citep{tsai2019multimodal} & Audio \& Video & - & 78.50* & -& -& -   \\ \midrule \multirow{8}{*}{\begin{tabular}[c]{@{}l@{}}Intermediate Attention \\  fusion \citep{9956592} \end{tabular}} 
					& Audio\textsuperscript{\textdaggerdbl} & - & 58.08*          & -            &-              & -              \\
					& Video\textsuperscript{\textdaggerdbl} & - & 72.83*         & -              & -                & -       \\   & Audio \& Video& \text{SUM + CE} & \underline{82.11\tiny{$\pm$4.57}}          & \underline{81.75\tiny{$\pm$4.43}}              & 81.77\tiny{$\pm$4.55}                & \underline{79.85\tiny{$\pm$5.18}}                  \\  & Audio \& Video& \text{NN + CE}\textsuperscript{\textdagger} & 80.29\tiny{$\pm$3.45}         & 79.77\tiny{$\pm$2.99}             & 80.02\tiny{$\pm$3.80}               & 77.78\tiny{$\pm$3.84}            \\
					& Audio \& Video& Mulmix             & 81.98\tiny{$\pm$2.38}            & 81.74\tiny{$\pm$2.08}                & \underline{81.85\tiny{$\pm$2.53}   }                 & 79.56\tiny{$\pm$2.70}                  \\
					& Audio \& Video& SoftTriple             & 71.78\tiny{$\pm$5.70}          & 71.44\tiny{$\pm$5.33}              & 71.34\tiny{$\pm$5.64}              & 68.20\tiny{$\pm$6.07}                \\
					& Audio \& Video& ProxyAnchor             & 71.08\tiny{$\pm$2.41}           & 70.47\tiny{$\pm$2.17}              & 71.19\tiny{$\pm$2.32}              & 67.32\tiny{$\pm$2.65}                \\
					
					& Audio \& Video & MultiModal                & \textbf{82.76\tiny{$\pm$4.60}} & \textbf{82.44\tiny{$\pm$4.50}}     & \textbf{82.55\tiny{$\pm$4.66}}        & \textbf{80.45\tiny{$\pm$5.27}}  \\\bottomrule
				\end{tabular}
			}
		\end{small}
	\end{center}
	\vskip -0.05in
	{\footnotesize \raggedright \par \hspace{4 em} * from \citep{9956592}.  \textsuperscript{\textdaggerdbl}{} is a performance in the presence of only audio and video modalities. \textsuperscript{\textdagger}{} is method in \citep{9956592}.\par}
	\vskip -0.2in
\end{table*}

\subsection{OPPORTUNITY Dataset}
We employed the OPPORTUNITY activity recognition dataset, which contains 19 modalities based on accelerometer and inertial unit data that support activity task recognition. Five classes of locomotion (\textit{i.e.}, stand, walk, sit, lie, and null) and 18 gestures (\textit{i.e.}, open/close door, open/close fridge, open/close dishwasher, open/close drawer, clean table, drink from cup, toggle switch, and null) were included.

{\bf Implementation Details.} 
We redesigned the DeepConvLSTM \citep{ordonez2016deep} model, which is an early fusion model that outperforms several other methods, as an late fusion model, and applied a loss function. We evaluated its performance on the OPPORTUNITY dataset as an early, intermediate, and late fusion types. DeepConvLSTM consists of four convolution layers, two long short-term memory (LSTM) layers, and a fully connected layer. 
The intermediate fusion model consists of four convolution layers for each of the 19 modalities, a concatenation layer, two LSTM layers, and one fully connected layer. The late fusion model consists of four convolution layers, two LSTM layers, and a fully connected layer per modality. We used the same hyperparameters for training, validation, and testing as in the DeepConvLSTM paper.

{\bf Results.}
Evidently from Table~\ref{table3}, NN+CE performed best on the relatively easy locomotion classification task, whereas  our loss performed second best. However, our loss outperformed on the more difficult gesture classification task.

\begin{table*}[ht]
	\vskip -0.15in
	\caption{Results on OPPORTUNITY for modes of locomotion and gesture classification.  }
	\label{table3}
	\vskip 0.1in
	\begin{center}
		\begin{small}
			\centering
			\setlength\tabcolsep{1.0pt}
			{\renewcommand{\arraystretch}{1.2}
				\begin{tabular}{lccccc|cccc}
					\toprule \multirow{2}{*}{{Fusion}} & \multirow{2}{*}{{Method}} &  \multicolumn{4}{c|}{{Locomotion Classification \scriptsize{(5 classes)}}}   & \multicolumn{4}{c}{{Gesture Classification \scriptsize{(18 classes)}}   }                        \\ \cmidrule{3-10} &   & {ACC}   & {FM}    & {FW}    & {MCC}   & {ACC}   & {FM}    & {FW}    & {MCC}  \\ \midrule
					Early\textsuperscript{\textdagger}  & -               & {86.00\tiny{$\pm$.39}}          & \text{ 87.53\tiny{$\pm$.37}}        & {85.77\tiny{$\pm$.42}}                & \text{{81.73\tiny{$\pm$.51}}}\hspace{0.2em}      & \text{ 88.84\tiny{$\pm$.73}}         &  58.51\tiny{$\pm$1.76}     & \underline{88.92\tiny{$\pm$.48}}           & \underline{63.87\tiny{$\pm$1.40}}      \\
					Intermed.    & -               & 84.35\tiny{$\pm$.90}          & 85.96\tiny{$\pm$.93}              & 84.10\tiny{$\pm$.96}                 & \text{79.56\tiny{$\pm$1.14}}\hspace{0.em}     & \text{ 85.89\tiny{$\pm$.59}}          & 41.82\tiny{$\pm$1.22}               & 85.16\tiny{$\pm$.56}                 & 50.06\tiny{$\pm$1.75}           \\ \midrule
					\multirow{5}{*}{Late }   & \text{SUM + CE}        & 85.90\tiny{$\pm$.23}    &  87.38\tiny{$\pm$.23}              & 85.73\tiny{$\pm$.24}           & 81.54\tiny{$\pm$.32}\hspace{0.2em}     & \text{ 88.90}\tiny{$\pm$.83}    & \underline{ 59.90\tiny{$\pm$1.13}}    & 88.87\tiny{$\pm$.52}                 & 63.55\tiny{$\pm$1.15}         \\
					& NN + CE           & \textbf{86.46\tiny{$\pm$.71}} & \textbf{87.97\tiny{$\pm$.55}}     & \textbf{86.28\tiny{$\pm$.68}}\hspace{0.2em}         & \textbf{82.29\tiny{$\pm$.95}} & \text{ 88.36}\tiny{$\pm$.89} & {57.84}\tiny{$\pm$.45}              & {88.40}\tiny{$\pm$.63}         & {62.25}\tiny{$\pm$1.69}  \\ & Mulmix        & - & -   & -       & - & - & -              &-         & -  \\ 
					& SoftTriple             & {85.97\tiny{$\pm$.41}} & {87.41\tiny{$\pm$.31}}     & 85.76\tiny{$\pm$.39}        & 81.66\tiny{$\pm$.55}\hspace{0.2em}  & \text{ 88.97}\tiny{$\pm$.46} & {58.68}\tiny{$\pm$1.03}             & {88.89}\tiny{$\pm$.35}        & {63.52}\tiny{$\pm$.96} \\ 
					& ProxyAnchor             & {85.77\tiny{$\pm$1.41}} & {87.40\tiny{$\pm$1.28}}     & {85.63\tiny{$\pm$1.48}}        & {81.38\tiny{$\pm$1.84}}\hspace{0.0em}  & \text{ \underline{89.56\tiny{$\pm$.58}}} & {53.76\tiny{$\pm$4.55}}             & {88.12\tiny{$\pm$1.04}}        & {60.65\tiny{$\pm$3.40}} \\ 
					& MultiModal             & \underline{86.11\tiny{$\pm$.67}} & \underline{87.57\tiny{$\pm$.60}}     & \underline{85.93\tiny{$\pm$.70}}        & \underline{81.84\tiny{$\pm$.84}}\hspace{0.2em}  & \textbf{\text{ 89.81\tiny{$\pm$.42}}} & \textbf{60.36\tiny{$\pm$1.81}}             & \textbf{89.48\tiny{$\pm$.45}}        & \textbf{65.01\tiny{$\pm$1.40}   } \\ \bottomrule
				\end{tabular}
			}
		\end{small}
	\end{center}
	\vskip -0.05in
	{\footnotesize \raggedright \par \hspace{3.5em} \textsuperscript{\textdagger}{} indicates a method used in \citep{ordonez2016deep}. MulMix requires $2^{19(\text{\# of modalities})}$$-$1 different mixtures of modalities, which \\ \hspace{3.7em} has a computational cost issue.  \par}
	\vskip -0.1 in
\end{table*}

\subsection{EPIC-KITCHENS Dataset}
We applied the loss function to the X-Norm model \citep{yin2022x}, a recent multimodal model on EPIC-KITCHENS-100, which is the largest public dataset for action recognition in first-person perspective. The dataset contains over 700,000 egocentric video clips at 59.94 frames per second. Each video is annotated with a caption containing one or more verbs and one or more nouns. For a fair comparison, we used the same hyperparameters and experimental setup as in the X-Norm paper. We used two modalities (RGB and computed optical flow datasets), with eight action classes (\textit{i.e.}, take, put, wash, open, close, insert, turn on and cut).

{\bf Results.} 
Table~\ref{table4} presents the results of applying our loss to the X-norm model on EPIC-KITCHENS. 
Evidently, our \text{MultiModal} loss exhibited better performance compared with the weighted SUM + CE (which is tuned to weights of 0.3 for RGB and 0.7 for optical flow) and other methods.

\vskip 0.05 in
\begin{table*}[t]
	\caption{Result on EPIC-KITCHENS for action recognition (8 classes).}
	\vskip -0.1 in
	\label{table4}
	\begin{center}
		\begin{small}
			\centering
			\setlength\tabcolsep{3.3pt}
			{\renewcommand{\arraystretch}{1.2}
				\begin{tabular}{lcccccc}
					\toprule
					\multicolumn{1}{l}{{Model}}& {Modality} & {Method}    & {ACC}   & {FM} & {FW} & {MCC}    \\ \midrule I3D \citep{carreira2017quo} & RGB  & - & 61.4* & -& -& -   \\ I3D \citep{carreira2017quo} & Optical flow& - & 60.0* & -& -& -   \\\midrule \multirow{7}{*}{X-Norm   \citep{yin2022x} } 
					& RGB \& Optical flow& \text{SUM + CE} & 71.11\tiny{$\pm$1.53} & 65.61\tiny{$\pm$1.85} & 70.81\tiny{$\pm$1.78} & 63.80\tiny{$\pm$1.94} \\  & RGB \& Optical flow& \text{Weighted SUM + CE}\textsuperscript{\textdagger} & \underline{71.26\tiny{$\pm$0.89}} & 65.91\tiny{$\pm$1.70}  & 71.05\tiny{$\pm$1.07} & \underline{64.34\tiny{$\pm$1.04}} \\  &RGB \& Optical flow& \text{NN + CE} & 71.24\tiny{$\pm$1.18}  & \textbf{66.48\tiny{$\pm$1.45}}  & \underline{71.12\tiny{$\pm$1.00}}                 & 64.02\tiny{$\pm$1.45}\\ & RGB \& Optical flow & Mulmix & 70.75\tiny{$\pm$0.84} & 64.63\tiny{$\pm$0.50} & 70.41\tiny{$\pm$0.64} & 63.37\tiny{$\pm$0.83} \\
					& RGB \& Optical flow& SoftTriple             & 70.36\tiny{$\pm$2.09} & 65.43\tiny{$\pm$1.22} & 70.62\tiny{$\pm$1.66} & 63.02\tiny{$\pm$2.42}\\
					& RGB \& Optical flow& ProxyAnchor             & 67.20\tiny{$\pm$1.73} & 63.07\tiny{$\pm$2.80} & 67.01\tiny{$\pm$1.84} & 58.60\tiny{$\pm$2.16}\\
					& RGB \& Optical flow & MultiModal              & \textbf{71.90\tiny{$\pm$0.46}} & \underline{66.24\tiny{$\pm$1.49}}     & \textbf{71.68\tiny{$\pm$0.79}}        & \textbf{64.84\tiny{$\pm$0.63}}  \\\bottomrule
				\end{tabular}
			}
		\end{small}
	\end{center}
	\vskip -0.05in
	{\footnotesize \raggedright \par \hspace{4em} * from \citep{yin2022x}. \textsuperscript{\textdagger}{} is a method in \citep{yin2022x} with weights of 0.3 for RGB and 0.7 for optical flow.\par}
	\vskip -0.05in
\end{table*}

\subsection{TCGA Dataset}
First, we applied the loss function to a recently developed  multimodal model for the TCGA cancer typing dataset, the multiomic multitask attention (MOMA) model \citep{moon2022moma}. MOMA is an intermediate two-step fusion model with a multitask learning step and an additional ensemble step that accounts for interactions between biological modalities through its attention mechanism. We used the preprocessed TCGA dataset, which contains 34 cancer types and 8,685 samples. Each sample contains gene expression (GE) and DNA methylation (DM) modalities. For a fair comparison, we used the reported MOMA settings, where we divided the dataset into five equal CVs and trained them with hyperparameters selected for each. 

{\bf Results.} 
Table~\ref{table5} presents the results of our loss compared with other fusion methods in terms of the accuracy (ACC), FW, area under the ROC curve (AUC), MCC, average precision (AP), and convergence time. Evidently, the traditional fusion methods (\textit{i.e.}, \text{SUM + CE}, \text{NN + CE}, and \text{Mulmix}) and proxy-based losses exhibited worse performance than the two-step ensemble approach of MOMA. \text{MultiModal} loss exhibited the best performance and converged approximately  six times faster than MOMA. Other proxy-based losses exhibited faster convergence compared with our loss.

\begin{table}[t]
	\caption{Results for TCGA cancer type classification (34 classes).}
	\vskip -0.12 in
	\label{table5}
	\begin{center}
		\begin{small}
			\centering
			\setlength\tabcolsep{3.2pt}
			{\renewcommand{\arraystretch}{1.15}
				\begin{tabular}{lcccccccc}
					\toprule
					{Model} & {Modality} & {Method} & {ACC}   & {FW}    & {AUC}   & {MCC}   & {AP}    & {Time (min)}   \\ \midrule
					
					\multirow{10}{*}{{\begin{tabular}[c]{@{}l@{}}MOMA \\  \citep{moon2022moma} \end{tabular}} }           & GE            & Multitask\textsuperscript{\textdagger}              & 94.99 {\scriptsize $\pm$.45}        & 94.74 \tiny{$\pm$.47}        & 99.62 \tiny{$\pm$.07}         & 94.76 \tiny{$\pm$.47}         & 97.65 \tiny{$\pm$.31}        & \multirow{3}{*}{85.2 \scriptsize{$\pm$7.0} } \\ 
					& DM            & Multitask\textsuperscript{\textdagger}              & 95.46 \tiny{$\pm$.30}          & 95.25 \tiny{$\pm$.31}         & 99.63 \tiny{$\pm$.11}         & 95.26 \tiny{$\pm$.31}         & 97.78 \tiny{$\pm$.26}         &                       \\ 
					& GE \& DM      & Ensemble\textsuperscript{\textdagger}        & \underline{95.84 \tiny{$\pm$.31} }         & \underline{95.72 \tiny{$\pm$.34}}          & \underline{99.64 \tiny{$\pm$.07}}          & \underline{95.65 \tiny{$\pm$.33}}         & \underline{97.96 \tiny{$\pm$.28}}    &                       \\ \cmidrule{2-9}
					& GE \& DM      & \text{SUM + CE}        & 94.38 \tiny{$\pm$.75}         & 93.38\tiny{$\pm$1.07}         & 99.47 \tiny{$\pm$.07}          & 94.14 \tiny{$\pm$.78}        & 97.58 \tiny{$\pm$.29}         & 92.7 \scriptsize{$\pm$4.6}                 \\
					& GE \& DM      & \text{NN + CE}         & 92.75 \tiny{$\pm$.30}         & 91.22 \tiny{$\pm$.43}         & 99.59 \tiny{$\pm$.09}         & 92.46 \tiny{$\pm$.31}         & 97.39 \tiny{$\pm$.39}         & 57.6 \scriptsize{$\pm$1.5}                 \\
					& GE \& DM      & \text{Mulmix}         & 94.92 \tiny{$\pm$.84}        & 94.51\tiny{$\pm$1.30}         & 99.06 \tiny{$\pm$.22}         & 94.70 \tiny{$\pm$.86}         & 96.93 \tiny{$\pm$.20}         & 53.9\scriptsize{$\pm$27.5}                 \\
					& GE \& DM      & \text{SoftTriple}         & 94.88 \tiny{$\pm$.43}        & 94.75 \tiny{$\pm$.41}         & 99.32 \tiny{$\pm$.31}         & 94.64 \tiny{$\pm$.45}         & 96.98 \tiny{$\pm$.20}         & \textbf{5.5 \scriptsize{$\pm$4.3}}                  \\												& GE \& DM      & \text{ProxyAnchor}         & 91.72\tiny{$\pm$1.06}        & 91.70\tiny{$\pm$1.11}          & 99.36\tiny{$\pm$.13}          & 91.46\tiny{$\pm$1.10}          & 97.85 \tiny{$\pm$.29}         & \underline{13.8} \scriptsize{$\pm$5.8}             \\
					& GE \& DM      & MultiModal              & \textbf{96.13 \tiny{$\pm$.16}} & \textbf{96.05 \tiny{$\pm$.16}} & \textbf{99.66 \tiny{$\pm$.14}}   & \textbf{95.95 \tiny{$\pm$.17}} & \textbf{98.01 \tiny{$\pm$.41}} & {15.1 \scriptsize{$\pm$0.5}}    \\ 
					\bottomrule
				\end{tabular}
			}
		\end{small}
	\end{center}
	\vskip -0.05in
	{\footnotesize \raggedright \par \hspace{4em}\textsuperscript{\textdagger}{} indicates methods used in \citep{moon2022moma}. \par}
\end{table}

\begin{figure}
	\vskip -0.05in
	\begin{floatrow}
		\CenterFloatBoxes
		\ffigbox[0.93\linewidth]{%
			\centerline{\includegraphics[width=0.47\textwidth]{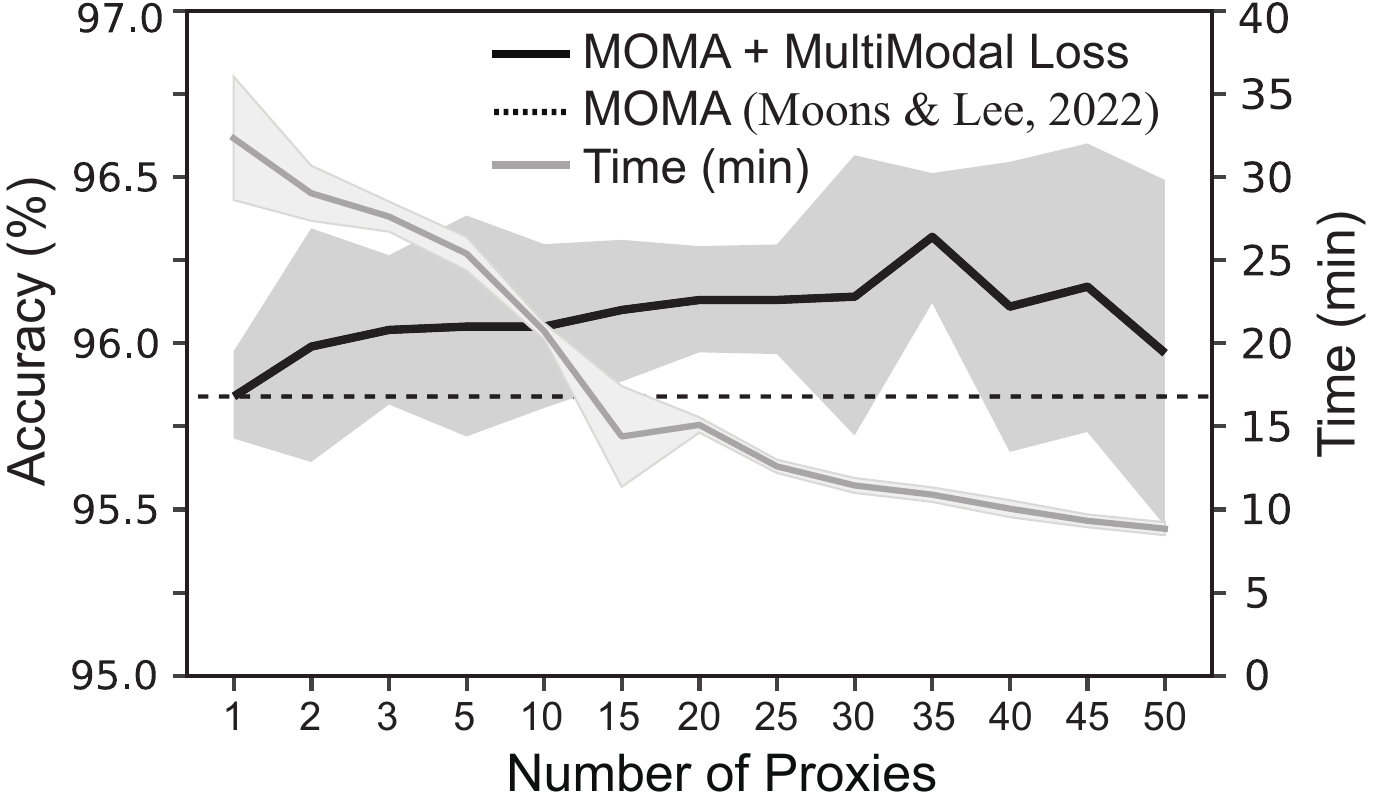}}
		}{\vskip -0.12in
			\caption{Accuracy and convergence time versus the number of proxies on the TCGA dataset. The standard deviation be shaded.} \label{fig4} 
		}
		\ttabbox[1.07\linewidth]{
			\begin{small}
				\centering
				\setlength\tabcolsep{1.75pt}
				{\renewcommand{\arraystretch}{1.4}
					\begin{tabular}{lcc}
						\toprule
						{Method} & {ACC}   & {Time (min)}   \\ \midrule
						
						\text{MultiModal (ours)}              & \textbf{96.13\tiny{$\pm$.16}} & \textbf{15.1\tiny{$\pm$0.5}}    \\ 		 \text{\hspace{0.7em}Soft Att.$\rightarrow$Hard Att.}              & {96.12}\tiny{$\pm$.21} &  {24.4\tiny{$\pm$2.3}}    \\ 			     \text{\hspace{0.7em}Norm. axis from class to proxy}           &  96.04\tiny{$\pm$.25}   & 18.4\tiny{$\pm$8.1}             \\   \text{\hspace{0.7em}w.o. Att.}           &  95.98\tiny{$\pm$.63}   & 15.9\tiny{$\pm$2.3}            \\  \makecell[l]{\hspace{0.5em} Norm. axis from class to proxy \\\hspace{0.5em} \& w.o. Att.}           &  95.03\tiny{$\pm$.40}          & 28.2\tiny{$\pm$1.0}                      \\ 
						\bottomrule
					\end{tabular}
				}
			\end{small}
		}{
			\caption{Ablation study for TCGA cancer type classification.}\label{table6}
		}
	\end{floatrow}
	\vskip -0.1in
\end{figure}

\section{Analysis}
\subsection{Impact of the Number of Proxies}
We investigated the effect of the number of proxies. Further, we evaluated and reportd the convergence time and the average ACC over five-fold CV on the TCGA dataset. The results of the analysis are quantified in Figure~\ref{fig4}. Evidently, as the number of proxies increased, the performance of our loss function improved and then deteriorated. When the number of proxies is large, the computational time can be reduced by subgrouping the outliers; however, this may lead to overfitting.

\subsection{Ablation study}
In Section 3, we introduced the component of \text{MultiModal} loss. To understand the importance of these components, we ablated and replaced them. We compared the convergence time and the average ACC over five-fold CV on TCGA dataset. Table~\ref{table6} presents the result; evidently, all the components contribute to the performance of \text{MultiModal} loss. Furthermore, we visualized the output of our full and pruned losses. We only summarize our results here owing to space constraints; details can be found in Appendix C. 

{\bf Replacing soft attention mechanisms with hard attention mechanisms.} The classification performance decreased slightly and the convergence time increased. This difference is owing to the different number of proxies considered. Soft attachments consider all proxies based on distance, whereas hard attachments consider only the closest proxy. 

{\bf Replacing the normalization axis from class to proxy.} Normalization on class aims to establish a one-to-one correspondence between the proxies of each class. Without normalization on class, performance was reduced. The number of proxies in the full model averages approximately 10 per class; however, in this case, the number of proxies was only 1--3 (Appendix C.2). 

{\bf Removing the attention mechanism.} The performance of our loss was reduced. The attention mechanism helps keep the output of each modality reliable (see Sec. 5.3 for details). Furthermore, the attention mechanism did not have a significant impact on the number of proxies (Appendix C.2). 

{\bf Replacing the normalization axis from class to proxy and removing the attention mechanism.} When we replaced the normalization axis from class to proxy and removed the attention mechanism, the classification performance and convergence speed deteriorated significantly. In addition, only one proxy per class was trained in this case (Appendix C.2).

\subsection{Learned Output Visualization} 
\begin{figure}
	\begin{center}
		\centerline{\includegraphics[width=1\textwidth]{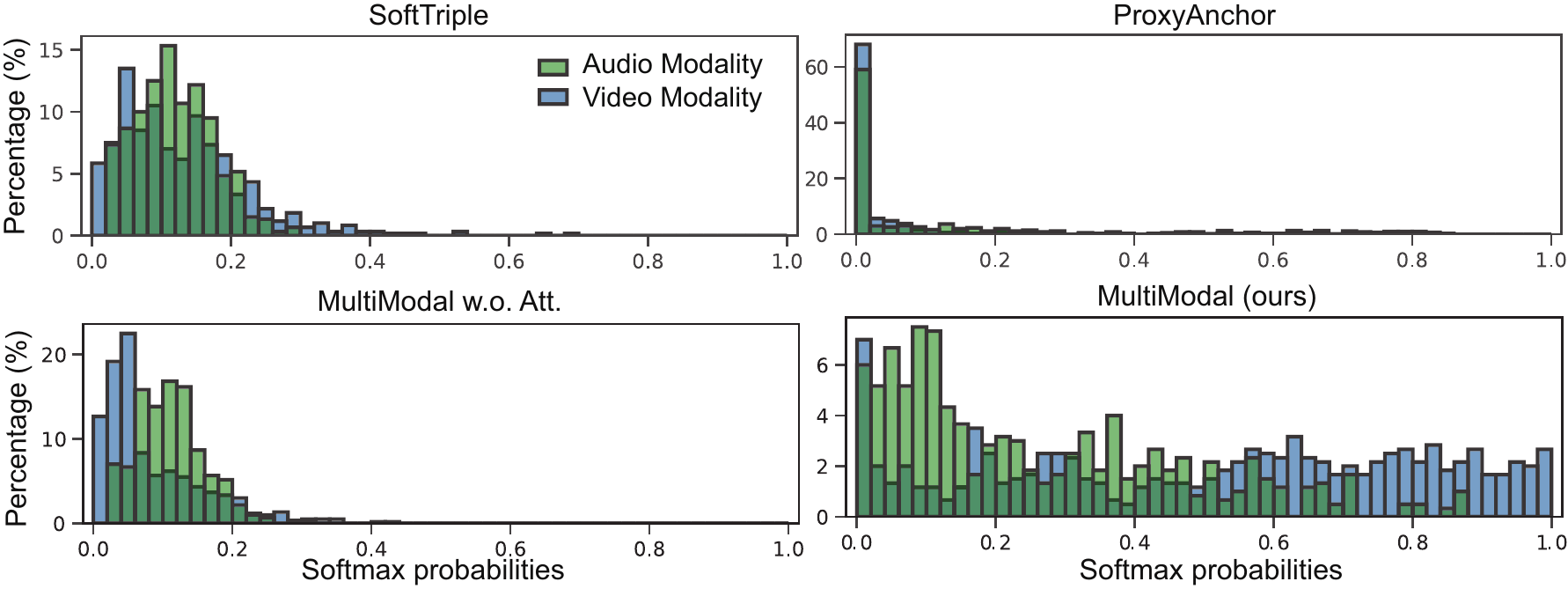}}
		\vskip -0.in
		\caption{Percentage histogram of predicted classification probabilities conditioned on true labels for each modality on RAVDESS.}
		\label{fig5}
	\end{center}
	\vskip -0.25in
\end{figure}

Finally, we analyzed the predicted classification probabilities for each modality using real-world data. The percentage histogram of the predicted classification probabilities conditioned on true labels by the Intermediate Attention Fusion model with \text{MultiModal} loss on RAVDESS is shown in Figure~\ref{fig5}. The existing proxy-based losses show results that do not consider the characteristics of each modality. By contrast, the output of our loss with the attention mechanism was more reliable and has a wider range of values. For \text{MultiModal} loss, the output of the video modality skewed toward a higher output probability compared with the audio modality. Interestingly, this is related to the observation that the video modality has higher unimodal predictive power than the audio modality \citep{9956592}. In Appendix C.1, we visualized the two-dimensional Uniform Manifold Approximation and Projection (UMAP) \citep{mcinnes2018umap} of the output from the two modalities using the MOMA model with our \text{MultiModal} loss on the TCGA dataset and found differences between different subgroups of a class. In Appendix C.3, we compared the visualization results of our loss with  alternative approaches (SUM + CE and SoftTriple) and found that using SUM+CE does not generate any diversity in the output within a class, and SoftTriple generates only one proxy per class. 

\section{Conclusions}
We presented a novel \text{MultiModal} loss function for subgrouping based on the contribution of each modality. The consistent improvements obtained by adding this loss to the benchmark multimodal learning models confirm its effectiveness. 
The results also indicate that the loss provides more reliable modality prediction and faster convergence. In addition, we validated our loss with analysis. We expect that this new loss function will inspire the development of more robust, explainable, and trustworthy multimodal models. 

{\bf Limitations.} A major limitation of our loss is that the multimodal model architecture must be an "intermediate" or "late" fusion model that has output from each modality. Additionally, although our loss works well for classification tasks, it is not applicable to regression tasks, which we aim to overcome in future research.

\bibliographystyle{unsrt}  
\bibliography{main} 

\begin{thebibliography}{10}

\bibitem{atrey2010multimodal}
Pradeep~K Atrey, M~Anwar Hossain, Abdulmotaleb El~Saddik, and Mohan~S
  Kankanhalli.
\newblock Multimodal fusion for multimedia analysis: a survey.
\newblock {\em Multimedia systems}, 16(6):345--379, 2010.

\bibitem{ramachandram2017deep}
Dhanesh Ramachandram and Graham~W Taylor.
\newblock Deep multimodal learning: A survey on recent advances and trends.
\newblock {\em IEEE signal processing magazine}, 34(6):96--108, 2017.

\bibitem{kamath2021mdetr}
Aishwarya Kamath, Mannat Singh, Yann LeCun, Gabriel Synnaeve, Ishan Misra, and
  Nicolas Carion.
\newblock Mdetr-modulated detection for end-to-end multi-modal understanding.
\newblock In {\em Proceedings of the IEEE/CVF International Conference on
  Computer Vision}, pages 1780--1790, 2021.

\bibitem{bendre2021generalized}
Nihar Bendre, Kevin Desai, and Peyman Najafirad.
\newblock Generalized zero-shot learning using multimodal variational
  auto-encoder with semantic concepts.
\newblock In {\em 2021 IEEE International Conference on Image Processing
  (ICIP)}, pages 1284--1288. IEEE, 2021.

\bibitem{gu2023dual}
Xin Gu, Yinghua Shen, and Chaohui Lv.
\newblock A dual-path cross-modal network for video-music retrieval.
\newblock {\em Sensors}, 23(2):805, 2023.

\bibitem{ghosal2018contextual}
Deepanway Ghosal, Md~Shad Akhtar, Dushyant Chauhan, Soujanya Poria, Asif Ekbal,
  and Pushpak Bhattacharyya.
\newblock Contextual inter-modal attention for multi-modal sentiment analysis.
\newblock In {\em proceedings of the 2018 conference on empirical methods in
  natural language processing}, pages 3454--3466, 2018.

\bibitem{xu2018co}
Nan Xu, Wenji Mao, and Guandan Chen.
\newblock A co-memory network for multimodal sentiment analysis.
\newblock In {\em The 41st international ACM SIGIR conference on research \&
  development in information retrieval}, pages 929--932, 2018.

\bibitem{wei2019neural}
Yinwei Wei, Xiang Wang, Weili Guan, Liqiang Nie, Zhouchen Lin, and Baoquan
  Chen.
\newblock Neural multimodal cooperative learning toward micro-video
  understanding.
\newblock {\em IEEE Transactions on Image Processing}, 29:1--14, 2019.

\bibitem{dong2019mlw}
Yunyun Dong, Wenkai Yang, Jiawen Wang, Juanjuan Zhao, Yan Qiang, Zijuan Zhao,
  Ntikurako Guy~Fernand Kazihise, Yanfen Cui, Xiaotong Yang, and Siyuan Liu.
\newblock Mlw-gcforest: a multi-weighted gcforest model towards the staging of
  lung adenocarcinoma based on multi-modal genetic data.
\newblock {\em BMC bioinformatics}, 20(1):1--14, 2019.

\bibitem{tsai2019multimodal}
Yao-Hung~Hubert Tsai, Shaojie Bai, Paul~Pu Liang, J~Zico Kolter, Louis-Philippe
  Morency, and Ruslan Salakhutdinov.
\newblock Multimodal transformer for unaligned multimodal language sequences.
\newblock In {\em Proceedings of the conference. Association for Computational
  Linguistics. Meeting}, volume 2019, page 6558. NIH Public Access, 2019.

\bibitem{su2020msaf}
Lang Su, Chuqing Hu, Guofa Li, and Dongpu Cao.
\newblock Msaf: Multimodal split attention fusion.
\newblock {\em arXiv preprint arXiv:2012.07175}, 2020.

\bibitem{joze2020mmtm}
Hamid Reza~Vaezi Joze, Amirreza Shaban, Michael~L Iuzzolino, and Kazuhito
  Koishida.
\newblock Mmtm: Multimodal transfer module for cnn fusion.
\newblock In {\em Proceedings of the IEEE/CVF Conference on Computer Vision and
  Pattern Recognition}, pages 13289--13299, 2020.

\bibitem{siddiqui2020multimodal}
Mohammad Faridul~Haque Siddiqui and Ahmad~Y Javaid.
\newblock A multimodal facial emotion recognition framework through the fusion
  of speech with visible and infrared images.
\newblock {\em Multimodal Technologies and Interaction}, 4(3):46, 2020.

\bibitem{delbrouck2020transformer}
Jean-Benoit Delbrouck, No{\'e} Tits, Mathilde Brousmiche, and St{\'e}phane
  Dupont.
\newblock A transformer-based joint-encoding for emotion recognition and
  sentiment analysis.
\newblock {\em arXiv preprint arXiv:2006.15955}, 2020.

\bibitem{sood2021multimodal}
Ekta Sood, Fabian K{\"o}gel, Philipp M{\"u}ller, Dominike Thomas, Mihai Bace,
  and Andreas Bulling.
\newblock Multimodal integration of human-like attention in visual question
  answering.
\newblock {\em arXiv preprint arXiv:2109.13139}, 2021.

\bibitem{saad2021learning}
Maliazurina Saad, Shenghua He, Wade Thorstad, Hiram Gay, Daniel Barnett, Yujie
  Zhao, Su~Ruan, Xiaowei Wang, and Hua Li.
\newblock Learning-based cancer treatment outcome prognosis using multimodal
  biomarkers.
\newblock {\em IEEE Transactions on Radiation and Plasma Medical Sciences},
  6(2):231--244, 2021.

\bibitem{chen2021multimodal}
Richard~J Chen, Ming~Y Lu, Wei-Hung Weng, Tiffany~Y Chen, Drew~FK Williamson,
  Trevor Manz, Maha Shady, and Faisal Mahmood.
\newblock Multimodal co-attention transformer for survival prediction in
  gigapixel whole slide images.
\newblock In {\em Proceedings of the IEEE/CVF International Conference on
  Computer Vision}, pages 4015--4025, 2021.

\bibitem{moon2022moma}
Sehwan Moon and Hyunju Lee.
\newblock Moma: a multi-task attention learning algorithm for multi-omics data
  interpretation and classification.
\newblock {\em Bioinformatics}, 38(8):2287--2296, 2022.

\bibitem{9956592}
Kateryna Chumachenko, Alexandros Iosifidis, and Moncef Gabbouj.
\newblock Self-attention fusion for audiovisual emotion recognition with
  incomplete data.
\newblock In {\em 2022 26th International Conference on Pattern Recognition
  (ICPR)}, pages 2822--2828, 2022.

\bibitem{wang2022fmfn}
Jingzi Wang, Hongyan Mao, and Hongwei Li.
\newblock Fmfn: Fine-grained multimodal fusion networks for fake news
  detection.
\newblock {\em Applied Sciences}, 12(3):1093, 2022.

\bibitem{wang2022energy}
Kang Wang, Youyi Song, Hongsheng Sheng, Jinghua Xu, Shuyou Zhang, and Jing Qin.
\newblock Energy efficiency design for eco-friendly additive manufacturing
  based on multimodal attention fusion.
\newblock {\em Journal of Manufacturing Processes}, 79:720--730, 2022.

\bibitem{yin2022x}
Yufeng Yin, Jiashu Xu, Tianxin Zu, and Mohammad Soleymani.
\newblock X-norm: Exchanging normalization parameters for bimodal fusion.
\newblock In {\em Proceedings of the 2022 International Conference on
  Multimodal Interaction}, pages 605--614, 2022.

\bibitem{liu2018learn}
Kuan Liu, Yanen Li, Ning Xu, and Prem Natarajan.
\newblock Learn to combine modalities in multimodal deep learning.
\newblock {\em arXiv preprint arXiv:1805.11730}, 2018.

\bibitem{wu2021spatiotemporal}
Hanbo Wu, Xin Ma, and Yibin Li.
\newblock Spatiotemporal multimodal learning with 3d cnns for video action
  recognition.
\newblock {\em IEEE Transactions on Circuits and Systems for Video Technology},
  32(3):1250--1261, 2021.

\bibitem{li2020sgm}
Jianan Li, Xuemei Xie, Qingzhe Pan, Yuhan Cao, Zhifu Zhao, and Guangming Shi.
\newblock Sgm-net: Skeleton-guided multimodal network for action recognition.
\newblock {\em Pattern Recognition}, 104:107356, 2020.

\bibitem{kashevnik2021multimodal}
Alexey Kashevnik, Igor Lashkov, Alexandr Axyonov, Denis Ivanko, Dmitry Ryumin,
  Artem Kolchin, and Alexey Karpov.
\newblock Multimodal corpus design for audio-visual speech recognition in
  vehicle cabin.
\newblock {\em IEEE Access}, 9:34986--35003, 2021.

\bibitem{song2022multimodal}
Qiya Song, Bin Sun, and Shutao Li.
\newblock Multimodal sparse transformer network for audio-visual speech
  recognition.
\newblock {\em IEEE Transactions on Neural Networks and Learning Systems},
  2022.

\bibitem{eddy2020integrated}
Sean Eddy, Laura~H Mariani, and Matthias Kretzler.
\newblock Integrated multi-omics approaches to improve classification of
  chronic kidney disease.
\newblock {\em Nature Reviews Nephrology}, 16(11):657--668, 2020.

\bibitem{moon2022sdgcca}
Sehwan Moon, Jeongyoung Hwang, and Hyunju Lee.
\newblock Sdgcca: Supervised deep generalized canonical correlation analysis
  for multi-omics integration.
\newblock {\em Journal of Computational Biology}, 29(8):892--907, 2022.

\bibitem{bromley1993signature}
Jane Bromley, Isabelle Guyon, Yann LeCun, Eduard S{\"a}ckinger, and Roopak
  Shah.
\newblock Signature verification using a" siamese" time delay neural network.
\newblock {\em Advances in neural information processing systems}, 6, 1993.

\bibitem{chopra2005learning}
Sumit Chopra, Raia Hadsell, and Yann LeCun.
\newblock Learning a similarity metric discriminatively, with application to
  face verification.
\newblock In {\em 2005 IEEE Computer Society Conference on Computer Vision and
  Pattern Recognition (CVPR'05)}, volume~1, pages 539--546. IEEE, 2005.

\bibitem{weinberger2009distance}
Kilian~Q Weinberger and Lawrence~K Saul.
\newblock Distance metric learning for large margin nearest neighbor
  classification.
\newblock {\em Journal of machine learning research}, 10(2), 2009.

\bibitem{wang2014learning}
Jiang Wang, Yang Song, Thomas Leung, Chuck Rosenberg, Jingbin Wang, James
  Philbin, Bo~Chen, and Ying Wu.
\newblock Learning fine-grained image similarity with deep ranking.
\newblock In {\em Proceedings of the IEEE conference on computer vision and
  pattern recognition}, pages 1386--1393, 2014.

\bibitem{movshovitz2017no}
Yair Movshovitz-Attias, Alexander Toshev, Thomas~K Leung, Sergey Ioffe, and
  Saurabh Singh.
\newblock No fuss distance metric learning using proxies.
\newblock In {\em Proceedings of the IEEE International Conference on Computer
  Vision}, pages 360--368, 2017.

\bibitem{kim2020proxy}
Sungyeon Kim, Dongwon Kim, Minsu Cho, and Suha Kwak.
\newblock Proxy anchor loss for deep metric learning.
\newblock In {\em Proceedings of the IEEE/CVF Conference on Computer Vision and
  Pattern Recognition}, pages 3238--3247, 2020.

\bibitem{qian2019softtriple}
Qi~Qian, Lei Shang, Baigui Sun, Juhua Hu, Hao Li, and Rong Jin.
\newblock Softtriple loss: Deep metric learning without triplet sampling.
\newblock In {\em Proceedings of the IEEE/CVF International Conference on
  Computer Vision}, pages 6450--6458, 2019.

\bibitem{livingstone2018ryerson}
Steven~R Livingstone and Frank~A Russo.
\newblock The ryerson audio-visual database of emotional speech and song
  (ravdess): A dynamic, multimodal set of facial and vocal expressions in north
  american english.
\newblock {\em PloS one}, 13(5):e0196391, 2018.

\bibitem{chavarriaga2013opportunity}
Ricardo Chavarriaga, Hesam Sagha, Alberto Calatroni, Sundara~Tejaswi Digumarti,
  Gerhard Tr{\"o}ster, Jos{\'e} del~R Mill{\'a}n, and Daniel Roggen.
\newblock The opportunity challenge: A benchmark database for on-body
  sensor-based activity recognition.
\newblock {\em Pattern Recognition Letters}, 34(15):2033--2042, 2013.

\bibitem{damen2020epic}
Dima Damen, Hazel Doughty, Giovanni~Maria Farinella, Sanja Fidler, Antonino
  Furnari, Evangelos Kazakos, Davide Moltisanti, Jonathan Munro, Toby Perrett,
  Will Price, et~al.
\newblock The epic-kitchens dataset: Collection, challenges and baselines.
\newblock {\em IEEE Transactions on Pattern Analysis and Machine Intelligence},
  43(11):4125--4141, 2020.

\bibitem{weinstein2013cancer}
John~N Weinstein, Eric~A Collisson, Gordon~B Mills, Kenna~R Shaw, Brad~A
  Ozenberger, Kyle Ellrott, Ilya Shmulevich, Chris Sander, and Joshua~M Stuart.
\newblock The cancer genome atlas pan-cancer analysis project.
\newblock {\em Nature genetics}, 45(10):1113--1120, 2013.

\bibitem{kingma2014adam}
Diederik~P Kingma and Jimmy Ba.
\newblock Adam: A method for stochastic optimization.
\newblock {\em arXiv preprint arXiv:1412.6980}, 2014.

\bibitem{luna2021proposal}
Cristina Luna-Jim{\'e}nez, Ricardo Kleinlein, David Griol, Zoraida Callejas,
  Juan~M Montero, and Fernando Fern{\'a}ndez-Mart{\'\i}nez.
\newblock A proposal for multimodal emotion recognition using aural
  transformers and action units on ravdess dataset.
\newblock {\em Applied Sciences}, 12(1):327, 2021.

\bibitem{ordonez2016deep}
Francisco~Javier Ord{\'o}{\~n}ez and Daniel Roggen.
\newblock Deep convolutional and lstm recurrent neural networks for multimodal
  wearable activity recognition.
\newblock {\em Sensors}, 16(1):115, 2016.

\bibitem{carreira2017quo}
Joao Carreira and Andrew Zisserman.
\newblock Quo vadis, action recognition? a new model and the kinetics dataset.
\newblock In {\em proceedings of the IEEE Conference on Computer Vision and
  Pattern Recognition}, pages 6299--6308, 2017.

\bibitem{mcinnes2018umap}
Leland McInnes, John Healy, and James Melville.
\newblock Umap: Uniform manifold approximation and projection for dimension
  reduction.
\newblock {\em arXiv preprint arXiv:1802.03426}, 2018.

\end{thebibliography}

\newpage
\appendix
\counterwithin{table}{section}
\counterwithin{figure}{section}
\renewcommand{\thetable}{\arabic{table}}
\renewcommand{\thefigure}{\arabic{figure}}

\textbf{\huge{Appendix}}

\def\thesection{\Alph{section}}

\section{Gradients respect to proxy and instance}
\subsection{MultiModal loss}
Detailing the gradient during training in terms of proxies and instances. After simplifying the formula in the main manuscript, we calculate the gradient.

\begin{equation}
	sim_{i,c}^k = \sum_{m}\left( {{x_{i}^{m}} \cdot w_c^{k,m}}\right).
\end{equation}
\begin{equation}
	att_{i,c}^k =sim_{i,c}^k.
\end{equation}
\begin{equation}
	A_{i,c} = \sum_{m}\left({(\sum_{c}{\sum_{k}att_{i,c}^k w_c^{k,m}+ E)}}\circ x_{i}^{m}\right).
\end{equation}
\begin{equation}
	S_{i,c}= \sum_{m} sim_{i,c}^m \cdot sim_{i,c}^m.
\end{equation}
\begin{equation}
	\ell_{\text{MultiModal}}  = -log{exp(S_{i,c}+A_{i,c})  \over \sum_{c} exp(S_{i,c}+A_{i,c})}. 
\end{equation}

We use the chain rule to compute the gradient for proxies and instances for normalised similarity scores $S_{i,c}$.
\begin{equation}
	\frac{\partial \ell_{\text{MultiModal}}}{\partial S_{i,c}}  = {{exp\left(S_{i,c} + A_{i,c} \right) \over \sum_{c}exp\left( {S_{i,c} + A_{i,c}}\right)}} - 1.
\end{equation}
\begin{equation}
	\dfrac{\partial S_{i,c}}{\partial sim_{i,c}^k} = 2sim_{i,c}^k,
	\frac{\partial sim_{i,c}^k}{\partial x_i^m} = w_{c}^{k,m}, \frac{\partial sim_{i,c}^k}{\partial  w_{c}^{k,m}} = x_i^m.
\end{equation}
The gradient of $\ell_{\text{MultiModal}}$ with respect to $x_i^m$ for $S_{i,c}$ is  computed with application of the chain rule.
\begin{equation}
	\begin{gathered}
		\frac{\partial \ell_{\text{MultiModal}}}{\partial x_i^m} \textit{ for $S_{i,c}$}= \sum_{c} \sum_{k} \frac{\partial \ell_{\text{MultiModal}}}{\partial S_{i,c}} \dfrac{\partial S_{i,c}}{\partial sim_{i,c}^k} \frac{\partial sim_{i,c}^k}{\partial x_i^m} 
		\\
		= \sum_{c} \sum_{k} \left({{exp\left(S_{i,c} + A_{i,c} \right) \over \sum_{c'}exp\left( {S_{i,c'} + A_{i,c'}}\right)}} - 1\right) \cdot 2sim_{i,c}^k\cdot w_{c}^{k,m}.
	\end{gathered}
\end{equation}
The gradient of $\ell_{\text{MultiModal}}$ with respect to $w_c^{k,m}$ for $S_{i,c}$ is  computed with application of the chain rule.
\begin{equation}
	\begin{gathered}
		\frac{\partial \ell_{\text{MultiModal}}}{\partial w_c^{k,m}}\textit{ for $S_{i,c}$}  =  \frac{\partial \ell_{\text{MultiModal}}}{\partial S_{i,c}} \dfrac{\partial S_{i,c}}{\partial sim_{i,c}^k} \frac{\partial sim_{i,c}^k}{\partial w_c^{k,m}} 
		\\
		=   \left({{exp\left(S_{i,c} + A_{i,c} \right) \over \sum_{c}exp\left( {S_{i,c} + A_{i,c}}\right)}} - 1\right) \cdot 2sim_{i,c}^k \cdot x_i^m.
	\end{gathered}
\end{equation}

Next, we compute the gradient to the proxy and instance via the attended output $A_{i,c}$.
\begin{equation}
	\frac{\partial \ell_{\text{MultiModal}}}{\partial A_{i,y}}  = {{exp\left(S_{i,y} + A_{i,y} \right) \over \sum_{c}exp\left( {S_{i,c} + A_{i,c}}\right)}} - 1.
\end{equation}
\begin{equation}
	\frac{\partial A_{i,y}}{\partial att_{i,c}^k} = \sum_{m}\left(w_{c,y}^{k,m}\circ x_{i,y}^{m}\right).
\end{equation}
\begin{equation}
	\frac{\partial att_{i,c}^k}{\partial sim_{i,c}^k}  = 1.
\end{equation}
\begin{equation}
	\frac{\partial A_{i,y}}{\partial x_{i,y}^{m}}  = \sum_{c}\sum_{k}att_{i,c}^{k}w_{c,y}^{k,m} + J.
\end{equation}
\begin{equation}
	\frac{\partial A_{i,y}}{\partial w_{c,y}^{k,m}}  =att_{i,c}^{k}x_{i,y}^{m}.
\end{equation}

The gradient of $\ell_{\text{MultiModal}}$ with respect to $x_{i,y}^m$ for $A_{i,y}$ is computed using the chain rule.
\begin{equation}
	\begin{gathered}
		\frac{\partial \ell_{\text{MultiModal}}}{\partial x_{i,y}^m} \textit{ for $ A_{i,y}$} = \frac{\partial \ell_{\text{MultiModal}}}{\partial A_{i,y}} \frac{\partial A_{i,y}}{\partial x_{i,y}^m} 
		\\
		= \left({{exp\left(S_{i,y} + A_{i,y} \right) \over \sum_{c}exp\left( {S_{i,c} + A_{i,c}}\right)}} - 1\right) \left(\sum_{c}\sum_{k}att_{i,c}^{k}w_{c,y}^{k,m} + 1\right).
	\end{gathered}
\end{equation}
The gradient of $\ell_{\text{MultiModal}}$ with respect to $w_{c,y}^{k,m}$ for $A_{i,y}$ is  computed with application of the chain rule.
\begin{equation}
	\begin{gathered}
		\frac{\partial \ell_{\text{MultiModal}}}{\partial w_{c,y}^{k,m}} \textit{ for $A_{i,y}$} =  \frac{\partial \ell_{\text{MultiModal}}}{\partial A_{i,y}} \frac{\partial A_{i,y}}{\partial w_{c,y}^{k,m}} 
		\\
		=    \left({{exp\left(S_{i,y} + A_{i,y} \right) \over \sum_{c}exp\left( {S_{i,c} + A_{i,c}}\right)}} - 1\right) \cdot att_{i,c}^{k} x_{i,y}^{m}.
	\end{gathered}
\end{equation}

Next, we compute the gradient to the proxy and instance via the attended output $A_{i,y}$.
The gradient of $\ell_{\text{MultiModal}}$ with respect to $x_i^m$ for $att_{i,c}^k$ is computed using the chain rule.
\begin{equation}
	\begin{gathered}
		\frac{\partial \ell_{\text{MultiModal}}}{\partial x_{i,y}^m} \textit{ for $att_{i,c}^k$} = \sum_{c} \sum_{k}\frac{\partial \ell_{\text{MultiModal}}}{\partial A_{i,y}} \frac{\partial A_{i,y}}{\partial att_{i,c}^k}\frac{\partial att_{i,c}^k}{\partial sim_{i,c}^k}  \frac{\partial sim_{i,c}^k}{\partial x_{i,y}^m} 
		\\
		= \sum_{c}\sum_{k} \left({{exp\left(S_{i,y} + A_{i,y} \right) \over \sum_{c'}exp\left( {S_{i,c'} + A_{i,c'}}\right)}} - 1 \right) \cdot \sum_{m'}\left(w_{c,y}^{k,m'}\circ x_{i,y}^{m'}\right) \cdot w_{c,y}^{k,m}.
	\end{gathered}
\end{equation}
The gradient of $\ell_{\text{MultiModal}}$ with respect to $w_{c,y}^{k,m}$ for $att_{i,c}^k$ is computed using the chain rule.
\begin{equation}
	\begin{gathered}
		\frac{\partial \ell_{\text{MultiModal}}}{\partial w_{c,y}^{k,m}} \textit{ for $att_{i,c}^k$} =  \frac{\partial \ell_{\text{MultiModal}}}{\partial A_{i,c}} \frac{\partial A_{i,c}}{\partial att_{i,c}^k}\frac{\partial att_{i,c}^k}{\partial sim_{i,c}^k}  \frac{\partial sim_{i,c}^k}{\partial w_{c,y}^{k,m}} 
		\\
		=    \left({{exp\left(S_{i,y} + A_{i,y} \right) \over \sum_{c'}exp\left( {S_{i,c'} + A_{i,c'}}\right)}} - 1 \right) \cdot \sum_{m'}\left(w_{c,y}^{k,m'}\circ x_{i,y}^{m'}\right) \cdot x_{i,y}^m.
	\end{gathered}
\end{equation}

\subsection{SoftTriple loss}
We calculate the gradient after simplifying the SoftTriple loss [2].

\begin{equation}
	\ell_{\text{SoftTriple}}  = -log{exp(S_{i,c})  \over \sum_{c} exp(S_{i,c})}. 
\end{equation}

\begin{equation}
	S_{i,c}= \sum_{k} sim_{i,c}^k \cdot sim_{i,c}^k.
\end{equation}
\begin{equation}
	sim_{i,c}^k = \left( {x_{i} \cdot w_c^{k}}\right).
\end{equation}

\begin{equation}
	\frac{\partial \ell_{\text{SoftTriple}}}{\partial S_{i,c}}    = {{exp\left(S_{i,c}  \right) \over \sum_{c}exp\left( {S_{i,c} }\right)}} - 1.
\end{equation}

\begin{equation}
	\dfrac{\partial S_{i,c}}{\partial sim_{i,c}^k} = 2sim_{i,c}^k,
	\frac{\partial sim_{i,c}^k}{\partial x_i} = w_{c}^{k}, \frac{\partial sim_{i,c}^k}{\partial  w_{c}^{k}} = x_i.
\end{equation}
The gradient of $\ell_{\text{SoftTriple}}$ with respect to $x_{i}$ is computed using the chain rule.
\begin{equation}
	\begin{gathered}
		\frac{\partial \ell_{\text{SoftTriple}}}{\partial x_i} =\sum_{c} \sum_{k}\frac{\partial \ell_{\text{SoftTriple}}}{\partial S_{i,c}} \dfrac{\partial S_{i,c}}{\partial sim_{i,c}^k} \frac{\partial sim_{i,c}^k}{\partial x_i} \\
		= \sum_{c} \sum_{k}\left({{exp\left(S_{i,c}  \right) \over \sum_{c'}exp\left( {S_{i,c'}}\right)}} - 1\right) \cdot 2sim_{i,c}^k  \cdot  w_c^k
	\end{gathered}
\end{equation}
The gradient of $\ell_{\text{SoftTriple}}$ with respect to $w_c^k$ is computed using the chain rule.
\begin{equation}
	\begin{gathered}
		\frac{\partial \ell_{\text{SoftTriple}}}{\partial w_c^k} = \frac{\partial \ell_{\text{SoftTriple}}}{\partial S_{i,c}} \dfrac{\partial S_{i,c}}{\partial sim_{i,c}^k} \frac{\partial sim_{i,c}^k}{\partial w_c^k}
		\\
		=  \left({{exp\left(S_{i,c}  \right) \over \sum_{c}exp\left( {S_{i,c}}\right)}} - 1\right) \cdot 2sim_{i,c}^k \cdot x_{i}.
	\end{gathered}
\end{equation}

\section{Hyperparameter tuning results on synthetic data}
The results of the hyperparameter tuning for each method are given in Table~\ref{A1}. We experimented with a training set and a validation set of synthetic data with a noise modality.  With a learning rate of 5e-5,  ProxyAnchor loss did not converge in 100,000 iterations.

\begin{table}[h]
	\caption{Performance on learning rate hyperparameters using validation set on synthetic data.}
	\vskip -0.1in
	\label{A1}
	\begin{center}
		\begin{small}
			\centering
			\setlength\tabcolsep{3.2pt}
			{\renewcommand{\arraystretch}{1.15}
				\begin{tabular}{lcccc}
					\toprule
					{Method} & {5e-3}   & {5e-4}    & {5e-5}   & {5e-6}   \\ \midrule
					
					Unimodal                   & \textbf{85.7 {\tiny $\pm$0.3}}       & 84.2 {\tiny $\pm$0.1}      & 84.1 {\tiny $\pm$0.0}          & 84.3 {\tiny $\pm$0.2} \\ 
					\text{SUM + CE}        & 81.3 \tiny{$\pm$1.7}         & \textbf{85.3 \tiny{$\pm$0.7}}         & 83.4 \tiny{$\pm$1.0}          & 82.5 \tiny{$\pm$0.6}            \\
					\text{NN + CE}         & 79.9 \tiny{$\pm$3.1}         & 80.1 \tiny{$\pm$2.3}         & \textbf{86.7 \tiny{$\pm$0.7}}        & 84.9 \tiny{$\pm$0.4}     \\
					\text{Mulmix}         & 80.3 \tiny{$\pm$2.1}        & \textbf{84.3 \tiny{$\pm$1.3}}         & 82.5 \tiny{$\pm$0.8}       & 82.3 \tiny{$\pm$1.1}                    \\
					\text{SoftTriple}          & 71.3\tiny{$\pm$15.2}         & \textbf{73.2\tiny{$\pm$17.2}}         & 70.9\tiny{$\pm$17.2}         & 66.9\tiny{$\pm$18.5}      \\												 \text{ProxyAnchor}         & 68.4\tiny{$\pm$15.8}        & 71.4\tiny{$\pm$15.0}          & \textbf{81.1 \tiny{$\pm$6.2}}          & \scriptsize{Did not converge}     \\
					MM (ours)              & 87.5 \tiny{$\pm$8.7} & \textbf{91.3 \tiny{$\pm$2.3}} & 90.5 \tiny{$\pm$2.9}   & 90.6 \tiny{$\pm$2.6}    \\ 
					\bottomrule
				\end{tabular}
			}
		\end{small}
	\end{center}
	\vskip -0.02in
	{\footnotesize \raggedright \par \hspace{13 em}  Bold indicates the learning rate we chose for each method. \par}
\end{table}

\section{Learned Output Visualization on TCGA data}

\subsection{Results of training with MultiModal loss}
\begin{figure}
	\begin{center}
		\centerline{\includegraphics[width=0.8\columnwidth]{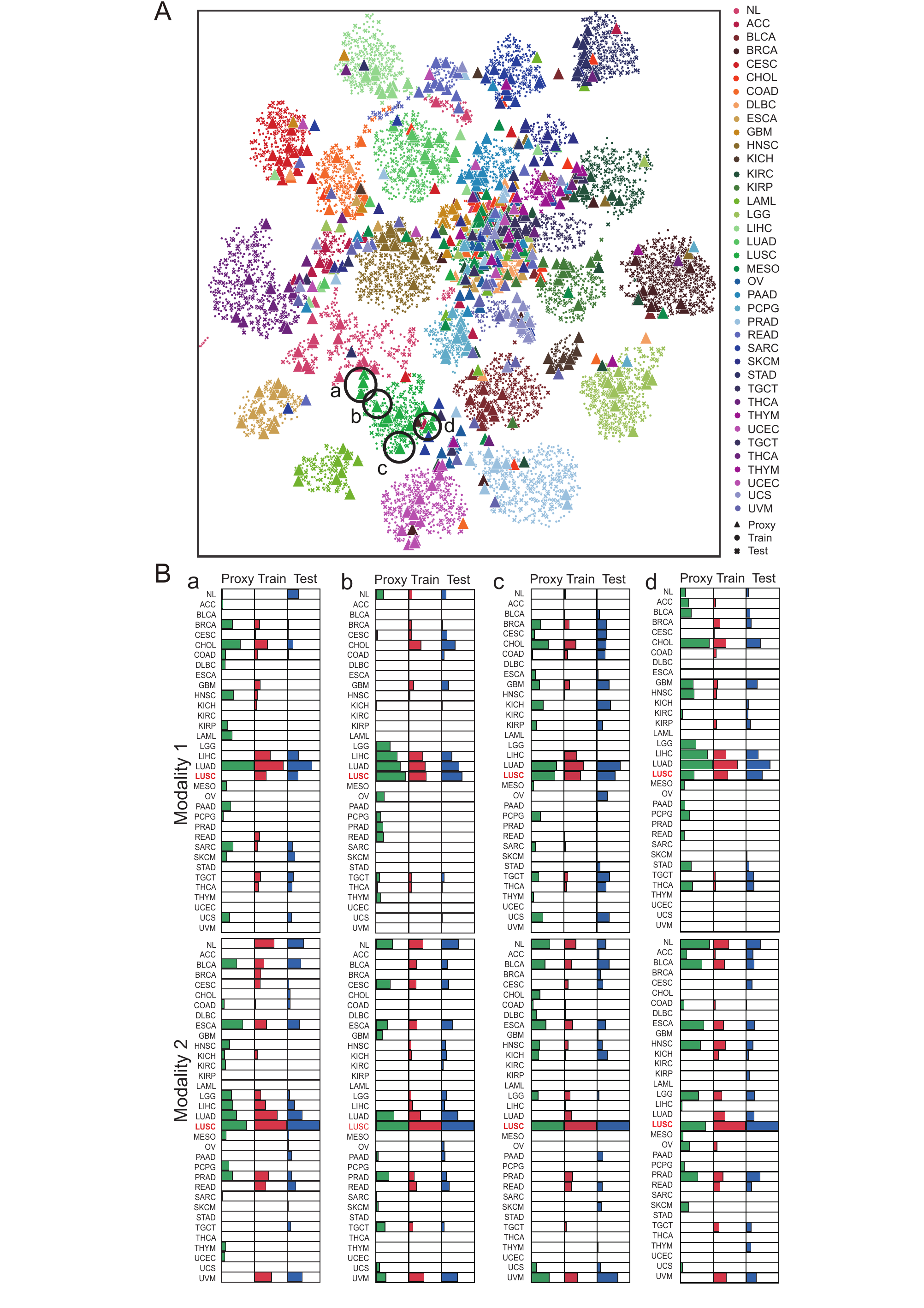}}
		\vskip -0.3in
		\caption{ (A) UMAP visualization of the two-dimensional embeddings of the learned outputs of each modality on TCGA is shown. (B) Examples of subgroups A, B, C and D for the LUSC class are shown, including proxies and instances. The size of the bars represents the relative probabilities of 34 cancer classes for four subgroups.}
		\label{A2}
	\end{center}
\end{figure}

To further investigate the subgrouping ability of our loss function, we visualized the two-dimensional Uniform Manifold Approximation and Projection (UMAP) [1] of the output learned from the two modalities using the MOMA model with our \text{MultiModal} loss on the TCGA dataset. Figure~\ref{A2} shows the proxies, training samples, and test samples for each class. We can see that even if we set the number of proxies to an excessive number, unused proxies are discarded in the middle. We further analysed the predicted probabilities across the sample subgroups within the lung squamous cell carcinoma (LUSC) class. We found that each subgroup showed different trends.  We found that each subgroup had a different tendency. For example, subgroup B had the highest probability of LUSC in both modalities, but subgroups A, C and D had the highest probability of lung adenocarcinoma (LUAD) in modality 1 and LUSC in modality 2. Interestingly, LUSC and LUAD are the two main subtypes of lung cancer. Among them, subgroup C had similar values with LUSC and LUAD in modality 1 and predicted LUSC with high probability in modality 2.

\subsection{Ablation study visualization}
\begin{figure}
	\begin{center}
		\centerline{\includegraphics[width=\columnwidth]{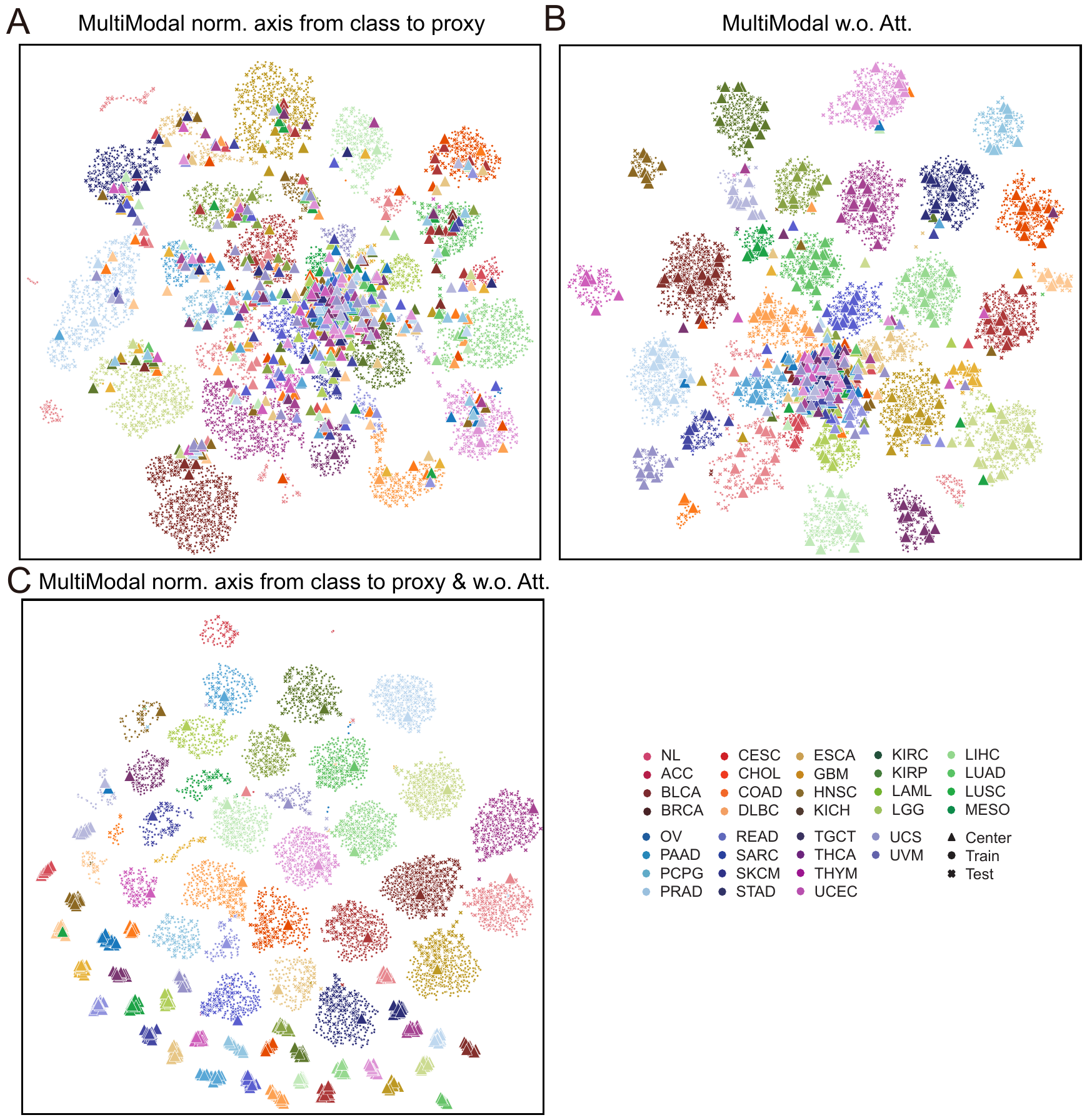}}
		
		\caption{UMAP visualization of the two-dimensional embeddings of the learned outputs from each modality on TCGA is shown. A, B, and C are the results of using MultiModal with normalization on proxy, MultiModal without Attention and MultiModal with normalization on proxy and Attention moved, respectively. }
		\label{A3}
	\end{center}
\end{figure}

In Figure~\ref{A3}, we have visualized the ablation study. When the normalization axis was changed from classes to proxies, the number of proxies was significantly reduced to 1-3. When we removed the attention mechanism, the number of proxies did not differ significantly. When we replaced the normalization axis with proxies in the class and removed the attention mechanism, there was only one proxy.

\subsection{Results of training with other methods}
\begin{figure}
	\begin{center}
		\centerline{\includegraphics[width=\columnwidth]{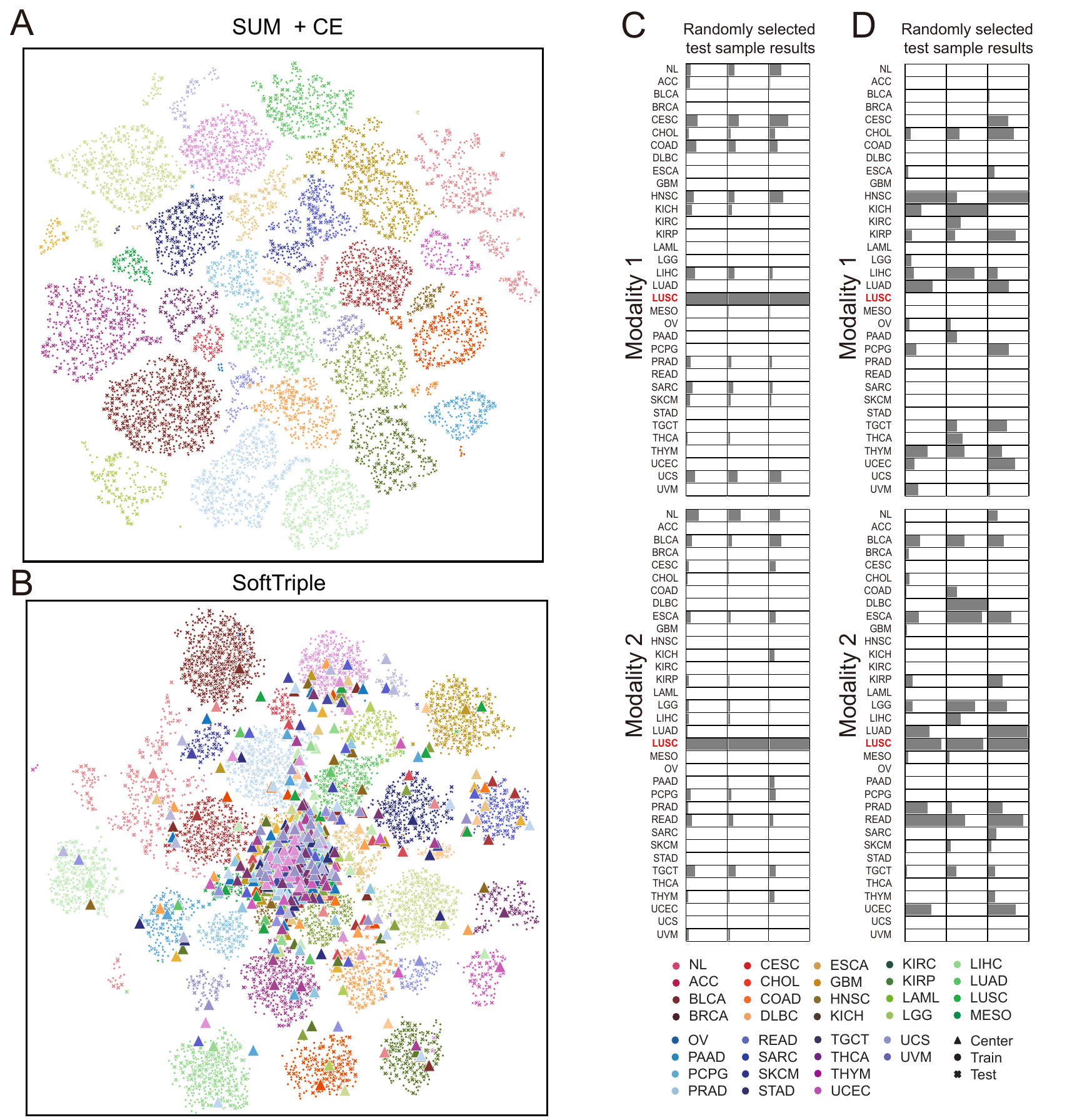}}
		
		\caption{UMAP visualization of the two-dimensional embeddings of the learned outputs from each modality on TCGA is shown. A and B are the results of using summation+CE and SoftTriple, respectively.  C and D show the outputs of a randomly selected test sample using summation+CE and SoftTriple, respectively. The size of the bars represents the relative probabilities of 34 cancer classes for four subgroups. }
		\label{A4}
	\end{center}
\end{figure}

In Figure~\ref{A4}, we have visualized the results learned with SUM+CE and SoftTriple. The results show that SUM+CE does not provide any diversity in the output of instances, while SoftTriple only generates only one proxy  per class.


\newpage
\textbf{References}

[1] Leland McInnes, John Healy, and James Melville. Umap: Uniform manifold approximation and projection for dimension reduction. \textit{arXiv preprint arXiv}:1802.03426, 2018. 

[2] Qi Qian, Lei Shang, Baigui Sun, Juhua Hu, Hao Li, and Rong Jin. Softtriple loss: Deep metric learning without
triplet sampling. In \textit{Proceedings of the IEEE/CVF International Conference on Computer Vision}, pages 6450–6458,
2019

\end{document}